\documentclass{article}

\usepackage[accepted]{icml2018}

\usepackage[utf8]{inputenc} % allow utf-8 input
\usepackage[T1]{fontenc}    % use 8-bit T1 fonts
\usepackage{url}            % simple URL typesetting
\usepackage{booktabs}       % professional-quality tables
\usepackage{amsfonts}       % blackboard math symbols
\usepackage{nicefrac}       % compact symbols for 1/2, etc.
\usepackage{microtype}      % microtypography2
\usepackage{multirow,array}
\usepackage{float}
\usepackage{algorithm}
\usepackage{algorithmic}

\usepackage{subfig}

\usepackage{graphicx} % more modern
\usepackage{amsmath}

\newtheorem{defn}{Definition}
\newtheorem{thm}{Theorem}

\usepackage{tikz}
\usetikzlibrary{arrows,positioning}
\tikzset{
    >=stealth,
    font=\scriptsize,
    possible world/.style={circle,draw,thick,align=center},
    real world/.style={double,circle,draw,thick,align=center},
    minimum size=60pt
}

\author{Adam Lerer\footnote{Facebook AI Research. Both authors contributed equally to this work and order was determined at random.} \\ Alexander Peysakhovich$^*$}

\begin{document}
\icmltitlerunning{Maintaining cooperation in complex social dilemmas using deep reinforcement learning}

\title{Maintaining cooperation in complex social dilemmas using deep reinforcement learning}

\maketitle

\begin{abstract}
Social dilemmas are situations where individuals face a temptation to increase their payoffs at a cost to total welfare. Building artificially intelligent agents that achieve good outcomes in these situations is important because many real world interactions include a tension between selfish interests and the welfare of others. We show how to modify modern reinforcement learning methods to construct agents that act in ways that are simple to understand, nice (begin by cooperating), provokable (try to avoid being exploited), and forgiving (try to return to mutual cooperation). We show both theoretically and experimentally that such agents can maintain cooperation in Markov social dilemmas. Our construction does not require training methods beyond a modification of self-play, thus if an environment is such that good strategies can be constructed in the zero-sum case (eg. Atari) then we can construct agents that solve social dilemmas in this environment. 
\end{abstract}

\section{Introduction}
Bilateral cooperative relationships, where individuals can pay costs to give larger benefits to others, are ubiquitous in our daily lives. In such situations mutual cooperation can lead to higher payoffs for all involved but there always exists an incentive to free ride. In a seminal work \citeauthor{axelrod2006evolution} asks a practical question: since social dilemmas are so ubiquitous, how should a person behave when confronted with one \cite{axelrod2006evolution}? In this work we will take up a variant of that question: how can we construct artificial agents that can solve complex bilateral social dilemmas? 

First, we must define what it means to `solve' a social dilemma. The simplest social dilemma is the two player, repeated Prisoner's Dilemma (PD). Here each player chooses to either cooperate or defect each turn. Mutual cooperation earns high rewards for both players. Defection improves one's payoff but only at a larger cost to one's partner. For the PD,  \citeauthor{axelrod1981evolution} suggest the strategy of tit-for-tat (TFT): begin by cooperating and in later turns copy whatever your partner did in the last turn \cite{axelrod1981evolution}. 

TFT and its variants (eg. Win-Stay-Lose-Shift, \cite{nowak1993strategy,imhof2007tit}) have been studied extensively across many domains including the social and behavioral sciences, biology, and computer science. TFT is popular for several reasons. First, it is able to avoid exploitation by defectors while reaping the benefits of cooperation with cooperators. Second, when TFT is paired with other conditionally cooperative strategies (eg. itself) it achieves cooperative payoffs. Third, it is error correcting because after an accidental defection is provides a way to return to cooperation. Fourth, it is simple to explain to a partner and creates good incentives: if one person commits to using TFT, their partner's best choice is to cooperate rather than try to cheat.

Our contribution is to expand the idea behind to TFT to a different environment: one shot Markov social dilemmas that require function approximation (eg. deep reinforcement learning). We will work with the standard deep RL setup: at training time, our agent is given access to the Markov social dilemma and can use RL to compute a strategy. At test time the agent is matched with an unknown partner and gets to play the game with that partner once. 

We will say that the agent can solve a social dilemma if it can satisfy the four TFT properties listed above. We call our strategy approximate (because we use RL function approximation) Markov (because the game is Markov) tit-for-tat (amTFT).

The first issue amTFT needs to tackle is that unlike in the PD `cooperation' and `defection' are no longer simple labeled strategies, but rather sequences of choices. amTFT uses modified self-play\footnote{We note that one advantage of amTFT is that it requires no additional machinery beyond what is required by standard self-play, thus if we can construct competitive agents in some environment (eg. Atari, \cite{mnih2015human}) then we can also construct agents that solve social dilemmas in that environment.} to learn two policies at training time: a fully cooperative policy and a `safe' policy (we refer to this as defection).\footnote{In the PD this action is `defect' but in most real social dilemmas this is not the case. For example social dilemmas occur naturally in economic situations where agents face a choice of trading and realizing gains from trade or simply producing everything they need on their own. In this case a safe policy is the outside option of `stop transacting with this agent.'}

The second issue is that we are considering a setup where our agent will only play the social dilemma once at test time. Thus the goal of amTFT is to intelligently switch between the learned policies within a single game.\footnote{The focus on a single test game is a way in which the problem we consider differs from what is normally studied in the literature on maintaining cooperation in repeated games \citep{fudenberg1986folk,littman2005polynomial,de2008polynomial}. In a standard `folk theorem' setup agents play a game repeatedly and maintain cooperation in one iteration of a game by threats of defection in the next iteration.} amTFT performs this as follows: at each time step during test time the amTFT agent computes the gain from the action their partner actually chose compared to the one prescribed by the cooperative policy. This can be done either using a learned $Q$ function or via policy rollouts. We refer to this as a per period debit. If the total debit is below a threshold amTFT behaves according to the cooperative policy. If the debit is above the threshold, the agent switches to the defecting policy for $k$ turns and then returns to cooperation. This $k$ is computed such that the partner's gains (debit) are smaller than the losses they incur ($k$ lost turns of cooperation). 

We show both analytically and experimentally that amTFT is a good strategy for Markov social dilemmas in the Axelrod sense defined above. Using a grid-world, Coins, and a modification of an Atari game where players must learn from pixels we demonstrate experimentally that an important component of amTFT is defining a partner's `defection' in terms of value and not actions. This choice makes amTFT robust to a partner using one of a class of outcome-equivalent cooperative policies as well function approximation, important properties for scaling agents beyond simple games.

We note that for the purposes of this paper we define the `cooperative' the policies as the ones which maximize the sum of both players' payoff. This definition seems natural for the case of the symmetric games we study (and is the one that is typically used in eg. the literature on the evolution of cooperation). However, it is well known that human social preferences take into account distribution (eg. inequity \cite{fehr1999theory}), various forms of altruism \citep{andreoni1990impure,peysakhovich2014humans}, and context dependent concerns (eg. social norms, see \cite{roth1991bargaining,herz2014makes,peysakhovich2015habits} for how social norms affect economic games and can change even in the lab). Thus when applying amTFT in other circumstances the correct `focal point' needs to be chosen. The automatic determination of focal points is an important topic for future research but far beyond the scope of this paper. However, we note that once this focal point is determined the amTFT algorithm can be used exactly as in this paper simply by swapping out the cooperative objective function during training time.

\subsection{Related Work}
A large literature on the `folk theorem' asks whether in a repeated game there exists an equilibrium which maintains cooperative payoffs using strategies which take as input histories of observations \citep{fudenberg1986folk,dutta1995folk} and output stage-game actions. A computer science branch of this literature asks whether it is possible to compute such equilibria either in repeated matrix games \citep{littman2005polynomial} or in repeated Markov games \citep{de2008polynomial}. These works are related to our questions but have two key differences: first, they focus on switching strategies \textit{across} iterations of a repeated game rather than within a single game. Second, perhaps more importantly, this literature focuses on finding equilibria unlike the Axelrod setup which focuses on finding a `good' strategy for a single agent. This difference in focus is starkly illustrated by TFT itself because both agents choosing TFT is \textbf{not} an equilibrium (since if one agent commits to TFT the partner's best response is not TFT, but rather always cooperate).

A second related literature focuses on learning and evolution in games \citep{fudenberg1998theory,sandholm1996multiagent,shoham2007if,nowak2006evolutionary,conitzer2007awesome} with recent examples applying deep learning to this question \citep{leibo2017multi,perolat2017multi}. Though there is a large component of this literature focusing on social dilemmas, these works typically are interested how properties of the environment (eg. initial states, payoffs, information, learning rules used) affect the final state of a set of agents that are governed by learning or evolutionary dynamics. This literature gives us many useful insights, but is not usually focused on the question of design of a single agent as we are.

A third literature focuses on situations where long term interactions with the same partner means that a good agent needs to either to discern a partner's type \citep{littman2001friend} or be able shape the adaptation of a learning partner \citep{babes2008social, foerster2017learning}. \citeauthor{babes2008social} use reward shaping in the Prisoner's Dilemma to construct `leader' agents that convince `followers' to cooperate and \citeauthor{foerster2017learning} uses a policy gradient learning rule which includes an explicit model of the partner's model. These works are related to ours but deal with situations where interactions are long enough for the partner to learn (rather than a single iteration) and require either explicit knowledge about the game structure \citep{babes2008social} or the partner's learning rule \citep{foerster2017learning}. 

There is a recent surge of interest in using deep RL to construct agents that can get high payoffs in multi-agent environments. Much of this literature focuses either on zero-sum environments \citep{tesauro1995temporal,silver2016mastering,silver2017mastering,brown2015hierarchical,kempka2016vizdoom,wu2016training,usunier2016episodic} or coordination games without an incentive to defect \citep{lowe2017multi,foerster2017stabilising,riedmiller2009reinforcement,tampuu2017multiagent,peysakhovich2017prosocial,lazaridou2017multi,das2017learning,evtimova2017emergent,havrylov2017emergence,foerster2016learning} and uses self-play to construct agents that can achieve good outcomes.\footnote{One example that is closer to our problem is \cite{lewis2017deal} which applies deep RL to bargaining, which can be thought of as an imperfect information general-sum game.} We show that applying this self-play approach naively does not produce agents that can solve social dilemmas (see the Appendix for more discussion).

There is a large literature using the repeated PD to study human decision-making in social dilemmas \citep{fudenberg2012slow,bo2011evolution}. In addition, recent work in cognitive science has begun to use more complex games and RL techniques quite related to ours \citep{weiner2016coordinate}. However, while this work provides useful insights into potentially useful strategies the main objective of this work is to understand human decision-making, not to actively improve the construction of agents.

Finally, \citeauthor{crandall2017cooperating} study how to construct machines for social dilemma games \cite{crandall2017cooperating}. This work is the closest to our own of the recent literature but differs in that it focuses specifically on cooperation with a human partner in simple games using cheap talk (English) communication from an exiting set of messages. Given the importance of communication in human interactions incorporating explicit signaling into amTFT-like strategies is an important and interesting direction for future work.

\section{The Basic Model}
We now turn to formalizing our main idea. We will work with a generalization of Markov decision problems:
\begin{defn}
A (finite, 2-player) Markov game consists of a set of \textit{states} $S = \lbrace s_1, \dots, s_n \rbrace$; a set of actions for each player $\mathcal{A}_1 = \lbrace a^1_1, \dots, a^1_k \rbrace$, $\mathcal{A}_2 = \lbrace a^2_1, \dots, a^2_k \rbrace$; a transition function $\tau: S \times A_1 \times A_2 \to \Delta(S)$ which tells us the probability distribution on the next state as a function of current state and actions; a reward function for each player $R_i : S \times A_1 \times A_2 \to \mathbb{R}$ which tells us the utility that player gains from a state, action tuple. We assume rewards are bounded.
\end{defn}

Players can choose between policies which are maps from states to probability distributions on actions $\pi_i: S \to \Delta (\mathcal{A}_i).$ We denote by $\Pi_i$ the set of all policies for a player. Through the course of the paper we will use the notation $\pi$ to refer to some abstract policy and $\hat{\pi}$ to learned approximations of it (eg. the output of a deep RL procedure).

\begin{defn}
A value function for a player $i$ inputs a state and a pair of policies $V^i(s, \pi_1, \pi_2)$ and gives the expected discounted reward to that player from starting in state $s$. We assume agents discount the future with rate $\delta$ which we subsume into the value function. A related object is the $Q$ function for a player $i$ inputs a state, action, and a pair of policies $Q^i(s, \pi_1, \pi_2)$ and gives the expected discounted reward to that player from starting in state $s$ taking action $a$ and then continuing according to $\pi_1, \pi_2$ afterwards. 
\end{defn}

We will be talking about strategic agents so we often refer to the concept of a best response:

\begin{defn}
A policy for agent $j$ denoted $\pi_j$ is a best response starting at state $s$ to a policy $\pi_i$ if for any $\pi'_j$ and any $s'$ along the trajectory generated by these policies we have $V^j(s', \pi_i, \pi_j) \geq V^j(s', \pi_i, \pi'_j).$ We denote the set of such best responses as $BR_j(\pi_i, s).$ If $\pi_j$ obeys the inequality above for \textit{any} choice of state $s$ we call it a perfect best response.
\end{defn}

The set of stable states in a game is the set of equilibria. We call a policy for player $1$ and a policy for player $2$ a Nash equilibrium if they are best responses to each other. We call them a Markov perfect equilibrium if they are perfect best responses.

We are interested in a special set of policies:

\begin{defn}
Cooperative Markov policies starting from state $s$ $(\pi_1^C, \pi_2^C)$ are those which, starting from state $s$, maximize $V^1(s, \pi_1, \pi_2) + V^2(s, \pi_1, \pi_2).$ We let the set of cooperative policies be denoted by $\Pi_i^C (c).$ Let the set of policies which are cooperative from any state be the set of perfectly cooperative policies.
\end{defn}

A social dilemma is a game where there are no cooperative policies which form equilibria. In other words, if one player commits to always cooperate, there is a way for their partner to exploit them and earn higher rewards at their expense. Note that in a social dilemma there may be policies which achieve the \textit{payoffs} of cooperative policies because they cooperate on the trajectory of play and prevent exploitation by threatening non-cooperation on states which are never reached by the trajectory.\footnote{An example of such a policy is Grim Trigger in the one-memory repeated PD. Grim Trigger cooperates if its partner has always cooperated in the past and defects otherwise. Thus two Grim players achieve the payoffs of two cooperators but do not use policies that cooperate at \textit{every} state.}

The state representation used plays an important role in determining whether equilibria which achieve cooperative payoffs exist. Specifically, a policy which rewards cooperation today with cooperation tomorrow must be able to remember whether cooperation happened yesterday. In both of our example games, Coins and the PPD, if the game is played from the pixels without memory maintaining cooperation is impossible. This is because the current state does not contain information about past behavior of one's partner. 

Thus, some memory is required to create policies which maintain cooperation. This memory can be learned (eg. an RNN) or it can be an explicitly designed summary statistic (our approach). However, adding memory does not remove equilibria where both players always defect, so adding memory does not imply that self-play will find policies that maintain cooperation \citep{foerster2017learning,sandholm1996multiagent}. In the appendix we show that even in the simplest situation, the one memory repeated PD, always defecting equilibria can be more robust attractors than ones which maintain cooperation. amTFT is designed to get around this problem by using modified self-play to explicitly construct the cooperative and cooperation maintaining strategies as well as then switching rule. We begin with the theory behind amTFT.

\section{Approximate Markov TFT}
We begin with a social dilemma where pure cooperators can be exploited. We aim to construct a simple meta-policy which incentivizes cooperation along the path of play by switching intelligently between policies in response to its partner.

We assume that cooperative polices are exchangeable. That is, for any pair $(\pi^C_1, \pi^C_2), (\pi'^C_1, \pi'^C_2) \in \Pi_i^C (s)$ any combination of the two (eg. $(\pi^C_1, \pi'^C_2)$ is also in $\Pi_i^C (s))$ and that all pairs give a unique distribution of the total rewards between the two players.

If policies are not exchangeable or can give different distributions of the total payoff then in addition to having a cooperation problem, we also have a coordination problem (ie. in which particular way should agents cooperate? how should gains from cooperation be split?). This is an important question, especially if we want our agents to interact with humans, and is related to the notion of choosing focal points in coordination/bargaining games. However, a complete solution is beyond the scope of this work and will often depend on contextual factors. See eg. \cite{schelling1980strategy,roth1991bargaining,weiner2016coordinate,peysakhovich2017prosocial} for more detailed discussion.

For the social dilemma to be solvable, there must be strategies with worse payoffs to both players. Consider an equilibrium $(\pi^D_1, \pi^D_2)$ which has worse payoffs for player $2$. We assume that $(\pi^D_1, \pi^D_2)$ is an equilibrium even if played for a finite time, which we call $\pi^D$-dominance. We use $\pi^D$-dominance to bound the payoffs of a partner during the execution of a punishment phase, thus it is a sufficient but not necessary condition. We discuss in the Appendix how this assumption can be relaxed. To define this formally, we first introduce the notation of a compound policy $\pi^{X_k Z}$ which is a policy that behaves according to $X$ for $k$ turns and then $Z$ afterwards.

\begin{defn}
We say a game is $\pi^D$ dominant (for player $2$) if for any $k$, any state $s$, and any policy $\pi_{A}$ we have $$V_2 (s, \pi_1^{D_k C},  \pi_2^{D_k C}) \geq V_2 (s, \pi_1^{D _k C}, \pi_2^{A_k C}).$$
\end{defn}

In theory, with access to $\pi^C, \pi^D$, their $Q$ functions, and no noise or function approximation, we can construct amTFT as follows. Suppose the amTFT agent plays as player $1$ (the reverse is symmetric).

At the start of the game the amTFT agent begins in phase $C$. If the phase is $C$ then the agent plays according to $\pi^C$. At each time step, if the agent is in a $C$ phase, the agent looks at the action $a_2$ chosen by their partner. The agent computes $$d = Q^2_{CC} (s, \pi_1^C(s), a_2) - Q^2_{CC} (s, \pi_1^C(s), \pi^C_2 (s)).$$ If $d > 0$ then starting at the next time step when state $s'$ is reached the agent enters into a $D$ phase where they choose according to $\pi^D$ for $k$ periods. $k$ is computed such that $$V_2 (s', \pi^{D_k C}_1, \pi^{D_k C}_2) - V_2 (s', \pi^{C}_1, \pi^{C}_2) > \alpha d.$$ Here $\alpha > 1$ controls how often an agent can be exploited by a pure defector. After this $k$ is over the agent returns to the $C$ phase. The amTFT strategy gives a nice guarantee:

\begin{thm}
Define $d^{*} = \max_{\mathcal{A}_2, s} (Q^2_{CC}(s, \pi^C_1(s), a) -  Q^2_{CC}(s, \pi_1^C(s), \pi_2^C(s))).$ If for any state $s$ we have that $V_2 (s, \pi_1^C, \pi^C_2) - V_2 (s, \pi_1^D, \pi_2^D) > \frac{d^{*}}{\delta}$ then if player $1$ is an amTFT agent, a fully omniscient player $2$ maximizes their payoffs by behaving according to $\pi^C_2$ when $1$ is in a $C$ phase and $\pi^D_2$ when $1$ is in a $D$-phase. Thus, if agents start in the $C$ phase and there is no noise, they cooperate forever. If they start in a $D$ phase, they eventually return to a $C$ phase.
\end{thm}

The proof is quite simple and we relegate it to the Appendix. However, we now see that amTFT has the desiderata we have asked for: it is easy to explain, it cooperates with a pure cooperator, it does not get completely exploited by a pure defector,\footnote{In an infinite length game amTFT will get exploited an infinite number of times as it tries to return to cooperation after each $D$ phase. One potential way to avoid this to avoid this is to increase $\alpha$ at each $D$ phase.} and incentivizes cooperation along the trajectory of play.\footnote{This makes amTFT subtly different from TFT. TFT requires one's partner to cooperate even during the $D$ phase for the system return to cooperation. By contrast, amTFT allows any action during the $D$ phase, this makes it similar to the rPD strategy of Win-Stay-Lose-Shift or Pavlov \citep{nowak1993strategy}.}

\section{Constructing an amTFT Agent}
We now use RL methods to construct the components required for amTFT by approximating the cooperative and defect policies as well as the switching policy. To construct the required policies we use self-play and two reward schedules: selfish and cooperative. 

In the selfish reward schedule each agent $i$ treats the other agent just as a part of their environment and tries to maximize their own reward. We assume that RL training converges and we call the converged policies under the selfish reward schedule $\hat{\pi}_i^D$ and the associated $Q$ function approximations $\hat{Q}^i_{DD}$. If policies converge with this training then $\hat{\pi}^D$ is a Markov equilibrium (up to function approximation).
 
In the cooperative reward schedule each agent gets rewards both from their own payoff and the rewards the other agent receives. That is, we modify the reward function so that it is $$R^{CC}_{i} (s, a_1, a_2) = R_{1}(s, a_1, a_2) + R_{2}(s, a_1, a_2).$$ We call the converged policy and value function approximations $\hat{\pi}_i^C$ and $\hat{Q}^i_{CC}.$ In this paper we are agnostic to which learning algorithm is used to compute policies. 

In general there can be convergence issues with selfish self-play \citep{fudenberg1998theory,conitzer2007awesome,papadimitriou2007complexity} while in the cooperative reward schedule the standard RL convergence guarantees apply. The latter is because cooperative training is equivalent to one super-agent controlling both players and trying to optimize for a single scalar reward.

With the value functions and policies in hand from the procedure above, we can construct an amTFT meta-policy. For the purposes of this construction, we consider agent $1$ as the amTFT agent (but everything is symmetric). The amTFT agent keeps a memory state $(W_t, b_t)$ which both start at $0$. 

The amTFT agent sees the action $a'$ of their partner at time $t$ and approximates the gain from this deviation as $D_{t} = \hat{Q}^2_{CC} (s, a^2_t) - \hat{Q}^2_{CC} (s, \pi^C_2(s)).$ To compute this debit we can either use learned $Q$ functions or we can simply use rollouts. 

The amTFT agent accumulates the total payoff balance of their partner as $W_t = W_{t-1} + D_{t}.$ If $W_t$ is below a fixed threshold $T$ the amTFT agent chooses actions according to $\pi^C.$ If $W_t$ crosses a threshold $T$ the mTFT agent uses rollouts to compute a $k$ such that the partner loses more from $\hat{\pi}^{D_k C}$ relative to cooperation than some constant $\alpha$ times the current debit. The hyperparameters $T$ and $\alpha$ trade off robustness to approximation error and noise. Raising $T$ allows for more approximation error in the calculation of the debit but relaxes the incentive constraints on the agent's partner. Raising $\alpha$ makes the cost of defection higher but makes false positives more costly. The algorithm is formalized below:

\begin{algorithm}
\caption{Approximate Markov Tit For Tat (for Agent 1)}
\begin{algorithmic}
\STATE \textbf{Input:} $\hat{\pi}^{C}, \hat{\pi}^D$ and their $\hat{Q}; \alpha, T$
\STATE $b \gets 0, W \gets 0$
\WHILE {$Game$}

\STATE $\hat{Q}$ comes from model or rollouts
\STATE $D \gets \hat{Q}^{2}_{CC} (s, a^2) - \hat{Q}^{2}_{CC} (s, \hat{\pi}_2^C (s))$

\IF {$b = 0$}
\STATE Choose $a \gets \hat{\pi}^C_1(s)$
\STATE $W = W + D$
\ENDIF

\IF {$b > 0$}
\STATE Choose $a \gets \hat{\pi}_1^D(s)$
\STATE $b = b - 1$
\ENDIF

\IF {$W > T$}
\STATE Compute $\hat{k}(s, \alpha W)$ using rollouts
\STATE $b = \hat{k}(s, \alpha T)$ 
\STATE $W = 0$
\ENDIF

\ENDWHILE
\end{algorithmic}
\end{algorithm}

A key component of the amTFT strategy is the computation of the per period debit $D_{t}.$ In our experiments we do this via use batched policy rollouts (a similar procedure is used to calculate the length of the $D$ phase, $k$). Each rollout is computed as follows: 

\begin{figure*}[!htb]
  \centering
  \subfloat[Coins]{{\includegraphics[scale=.23]{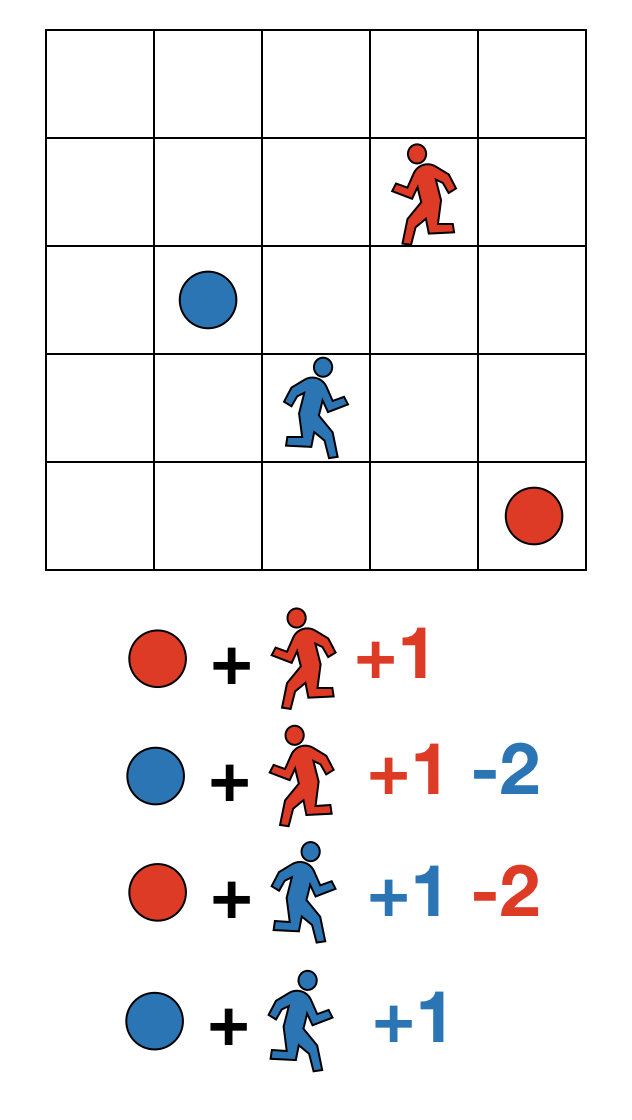}}} 
    \subfloat[PPD]{\raisebox{0cm}{\includegraphics[scale=.38]{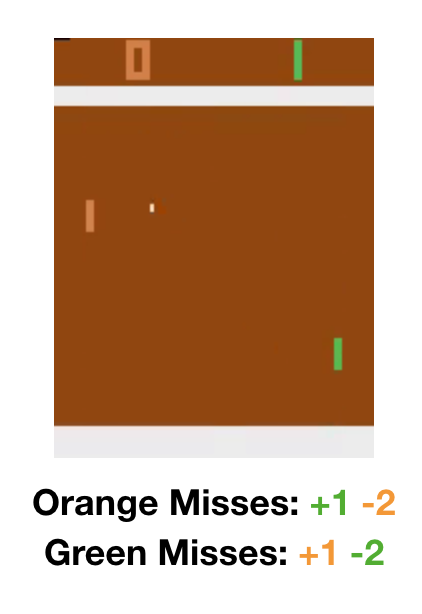}}}
  \subfloat[Training]{\includegraphics[scale=.42]{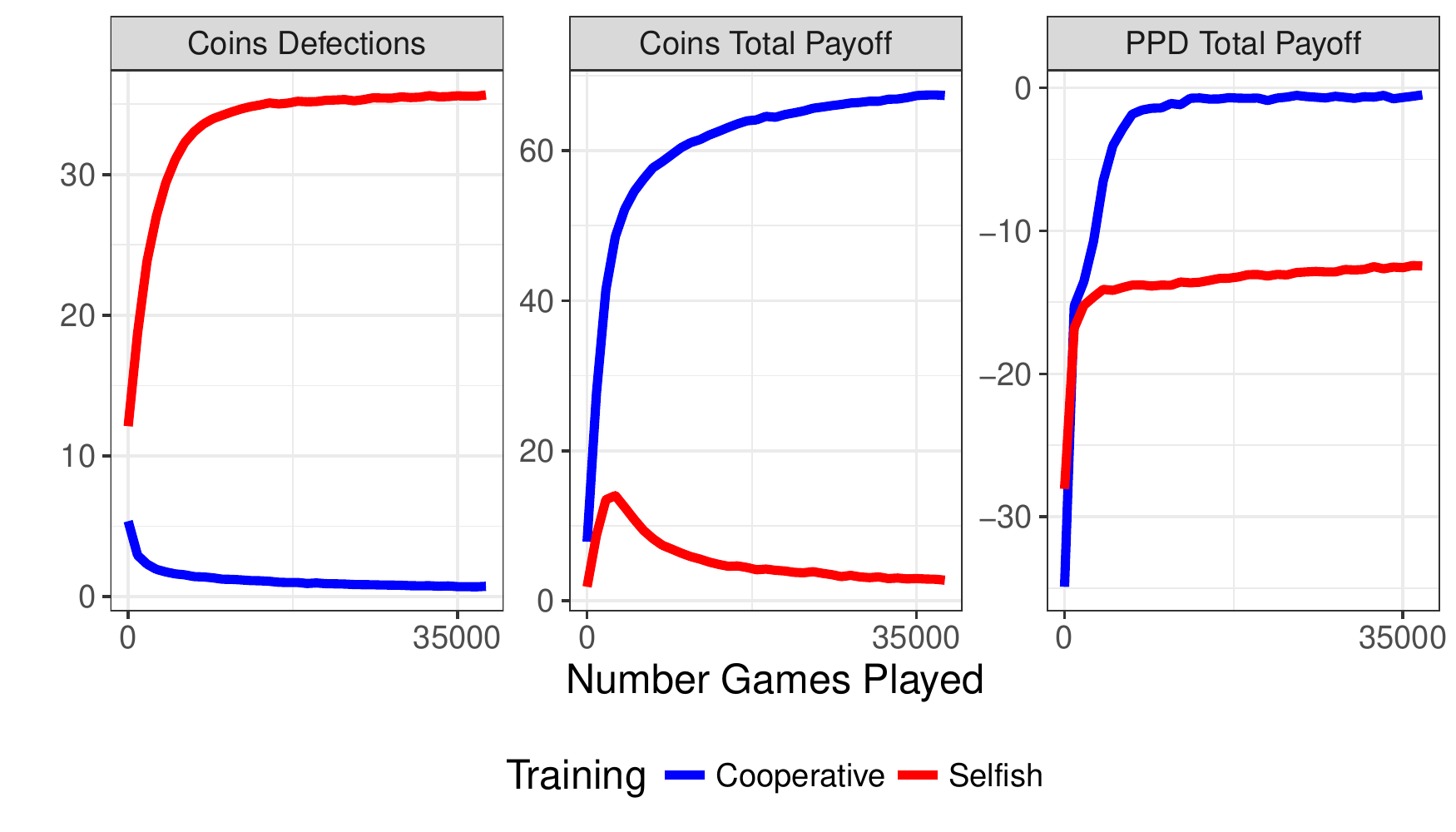}}
  \caption{In two Markov social dilemmas we find that standard self-play converges to defecting strategies while modified self-play finds cooperative, but exploitable strategies. We use the results of these two training schedules to construct $\hat{\pi}^C$ and $\hat{\pi}^D$.}
  \label{expt_results}
\end{figure*}

\begin{enumerate}
\item The amTFT agent has policy pairs $(\hat{\pi}_1^C, \hat{\pi}_2^C)$ and  $(\hat{\pi}_1^D, \hat{\pi}_2^D)$ saved from training
\item At time $t$ when the state was $s$ the amTFT agent compares the action chosen by their partner which we denote as $a'$ to $\hat{\pi}_2^C (s)$
\item If $a' = \hat{\pi}_2^C (s)$ then $D_t$ is set to $0$
\item If $a' \neq \hat{\pi}_2^C (s)$ then the amTFT agent simulates $2B$ replicas of the game for $M$ turns. 
\begin{enumerate}
\item In $B$ of the replicates their partner starts with $a'$ and play continues according to $(\hat{\pi}_1^C, \hat{\pi}_2^C)$ - we call this the `true path'
\item In $B$ of the replicates their partner starts with $\hat{\pi}_2^C(s)$ and play continues according to $(\hat{\pi}_1^C, \hat{\pi}_2^C)$ - we call this the `counterfactual path'
\item The amTFT agent takes the difference in the average total reward to the partner from the two paths and uses that as $D_t$ - this is an estimate of the reward of the one shot deviation to $a'$ from the recommended strategy $\hat{\pi}_2^C(s)$
\end{enumerate}
\end{enumerate}

This procedure is an unbiased estimator of $Q_{CC}$ in the limit of large $B$ and $M$ but is computationally intensive at test time.\footnote{A less computationally demanding way to execute amTFT is to use a model to approximate $Q_{CC}$ directly. This is difficult in practice since any bias in the model is accumulated across periods and because the model needs to be accurate everywhere, not just on the trajectory of $\pi^C$. In the appendix we discuss some results on learning a model $\hat{Q}$ and improving the efficiency of such procedures is an important direction for future work.} In games where an action today can only affect payoffs up to $M$ periods from now it suffices to use rollouts of length $M$ and elide the continuation value. 

The value-based construction gives amTFT a particular robustness property - if the partner is not using $\hat{\pi}_2^C$ exactly but is using a policy that is outcome equivalent to it the estimated $D_t$ values will end up being $0$ in expectation and so the amTFT agent will continue to cooperate. We will see in our experiments that this property is important to the success of amTFT in real Markov social dilemmas.

\section{Experiments}
\begin{figure*}
  \centering
  \subfloat[Coins]{

  \begin{tabular}{ l || c | c | c  } 
 \toprule
 Strategy & SelfMatch    & Safety & IncentC  \\
%& & & {\tiny $ S_2(X, C)-$} \\
%& {\tiny $S_1(X, X)$ } & {\tiny $S_1(X, D)$ } & {\tiny $S_2(X,D){\text\ \ \ \ }$}  \\
%Strategy & Nice & Safe & IncentC \\ [0.5ex] 

 \midrule
 $\pi^C$  & 68      & -58      & -41 \\ 
 $\pi^D$  & 2        & 0     & -58 \\
  amTFT  & 63     & -16 & 33 \\
  Grim      & 2       & 0  & -59 \\
 \bottomrule
 \end{tabular}
  }
 \subfloat[PPD]{

 \begin{tabular}{ l || c | c | c } 
 \toprule
 Strategy & SelfMatch    & Safety & IncentC  \\
%& & & {\tiny $ S_2(X, C)-$} \\
%& {\tiny $S_1(X, X)$ } & {\tiny $S_1(X, D)$ } & {\tiny $S_2(X,D){\text\ \ \ \ }$}  \\
\midrule
 $\pi^C$  & -1 & -18 & -14 \\ 
 $\pi^D$  & -7 & 0 & -19 \\
  amTFT  & -2  &  -1 & 2.5 \\
  Grim      & -7 & 0 & -18 \\
 \bottomrule
 \end{tabular}
 }
  \label{measures}
   \caption{In two Markov social dilemmas, amTFT satisfies the Axelrod desiderata: it mostly cooperates with itself, is robust against defectors, and incentivizes cooperation from its partner. The `Grim' strategy based on \cite{de2008polynomial} behaves almost identically to pure defection in these social dilemmas. The result of standard self-play is $\pi^D.$ The full tournament of all strategies against each other is shown in the Appendix.}
  \label{expt_results2}
\end{figure*}

We test amTFT in two environments: one grid-world and one where agents must learn from raw pixels. In the grid-world game Coins two players move on a $5 \times 5$ board. The game has a small probability of ending in every time step, we set this so the average game length is $500$ time steps. Coins of different colors appear on the board periodically, and a player receives a reward of 1 for collecting (moving over) any coin. However, if a player picks up a coin of the other player's color, the other player loses 2 points. The payoff for each agent at the end of each game is just their own point total. The strategy which maximizes total payoff is for each player to only pick up coins of their own color; however each player is tempted to pick up the coins of the other player's color. 

We also look at an environment where strategies must be learned from raw pixels. We use the method of \cite{tampuu2017multiagent} to alter the reward structure of Atari Pong so that whenever an agent scores a point they receive a reward of $1$ and the other player receives $-2$. We refer to this game as the Pong Player's Dilemma (PPD). In the PPD the only (jointly) winning move is not to play. However, a fully cooperative agent can be exploited by a defector.

We are interested in constructing general strategies which scale beyond tabular games so we use deep neural networks for state representation for both setups. We use standard setups so we relegate the details of the networks as well as the training to the appendix. 

We perform both Selfish (self play with reactive agents receiving own rewards) and Cooperative (self play with both agents receiving sum of rewards) training for both games. We train $100$ replicates for Coins and $18$ replicates for the PPD. In both games Selfish training leads to suboptimal behavior while Cooperative training does find policies that implement socially optimal outcomes. In Coins $\hat{\pi}^D$ agents converge to picking up coins of all colors while social $\hat{\pi}^C$ agents learn to only pick up matching coins. In PPD selfishly trained agents learn to compete and try to score while prosocially trained agents gently hit the ball back and forth.

We evaluate the performance of various Markov social dilemma strategies in a tournament. To construct a matchup between two strategies we construct agents and have them play a fixed length iteration of the game. Note that at training time we use a random length game but at test time we use a fixed length one so that we can compare payoffs more efficiently. We use $1000$ replicates per strategy pair to compute the average expected payoff. We compare $\hat{\pi}^C$, $\hat{\pi}^D$, and amTFT.

We also compare the direct adaptation of the construction in \cite{de2008polynomial}. Recall that the folk theorem algorithm maintains equilibria by threat of deviation later: if either agent's behavior in game iteration $t$ does not accord with the cooperative policy, both agents switch to a different policy in the next repetition of the game. We adapt this to the single test game setting as follows: the agent computes policies $\hat{\pi}^C, \hat{\pi}^D.$ If their partner $j$ takes an action $a$ in a state $s$ where $a \neq \hat{\pi}^C_j (s)$ the agent switches to $\hat{\pi}^D$ forever. We call this the Grim Trigger Strategy due to its resemblance to the rPD strategy of the same name.

In both games we ask how well the strategies satisfy Axelrod's desiderata from the introduction. Specifically, we would like to measure whether a strategy avoids exploitation, cooperates with conditional cooperators, and incentivizes its partner to cooperate. 

Let $S_i(X, Y)$ be the average reward to player $i$ when a policy of type $X$ is matched with type $Y$. The metric $$\text{Safety}(X) = S_1(X, D) - S_1(D, D).$$ measures how safe a strategy is from exploitation by a defector. The more negative this value, the worse that $\pi^X$ is exploited by a pure defector.

We measure a strategy's ability to achieve cooperative outcomes with policies of their same type as $$\text{SelfMatch}(X) = S_1(X, X).$$ This measure can be thought of as quantifying two things. First, how much social welfare is achieved in a world where \textit{everyone} behaves according to strategy $X$. Second, while we cannot enumerate all possible conditionally cooperative strategies, in the case of Grim and amTFT this serves as an indicator of how well they would behave against a particular conditional cooperator - themselves.

Finally, we measure if $X$ incentivizes cooperation from its partner. For this we use the measure $$\text{IncentC}(X) = S_2(X, C) - S_2(X, D).$$ The higher this number, the better off a partner is from committing to pure cooperation rather than trying to cheat.

Figure \ref{expt_results2} shows our metrics evaluated for the strategies of always cooperate, always defect, amTFT and Grim. Pure cooperation is fully exploitable and pure defection gets poor payoffs when matched with itself. Neither pure C or pure D incentivizes cooperation. However, amTFT avoids being exploited by defectors, does well when paired with itself and incentivizes cooperative strategies from its partner. We also see that inferring a partner's cooperation using the value function (amTFT) is much more stable than inferring it via actions as Grim immediately interprets any deviation from its preferred cooperative strategy as defection.

\section{amTFT As Teacher}
The results above show that amTFT is a good strategy to employ in a mixed environment which includes some cooperators, some tit-for-tat agents and some defectors. In particular, we have shown that amTFT is not exploited by $\pi^D$. However, what happens when amTFT's partner is themselves a learning agent? 

We consider what happens if we fix the one player (the Teacher) to use a fixed policy but let the other player be a selfish deep RL agent (the Learner). We perform the retraining in the domain of Coins.\footnote{We tried to perform the retraining in the PPD but incentivizing cooperation via a shift to $\pi^D$ requires a low discount rate and we found A3C to be unstable in this regime.} This retraining procedure can also be used as an additional metric of the exploitability of a given strategy, rather than asking whether $\hat{\pi}^D$ can exploit it, we ask whether a learner trying to maximize its own payoff can find some way to cheat.

Recall that when selfish RL agents played with each other, they converged to the Selfish `grab all coins' strategy. We see that Learners paired with purely cooperative teachers learn to exploit the teachers, learners paired with $\hat{\pi}^D$ also learn to exploit (this learning happens much slower because a fully trained $\hat{\pi}^D$ policy is able to grab coins very quickly and thus it is hard for a blank slate agent to learn at all), however learners paired with amTFT learn to cooperate. Note that choosing amTFT as a strategy leads to higher payoffs for both the Learner and the Teacher, thus even if we only care about the payoffs accrued to our own agent we can do better with amTFT than a purely greedy strategy.

\begin{figure}[h]
\centering
\includegraphics[scale=.5]{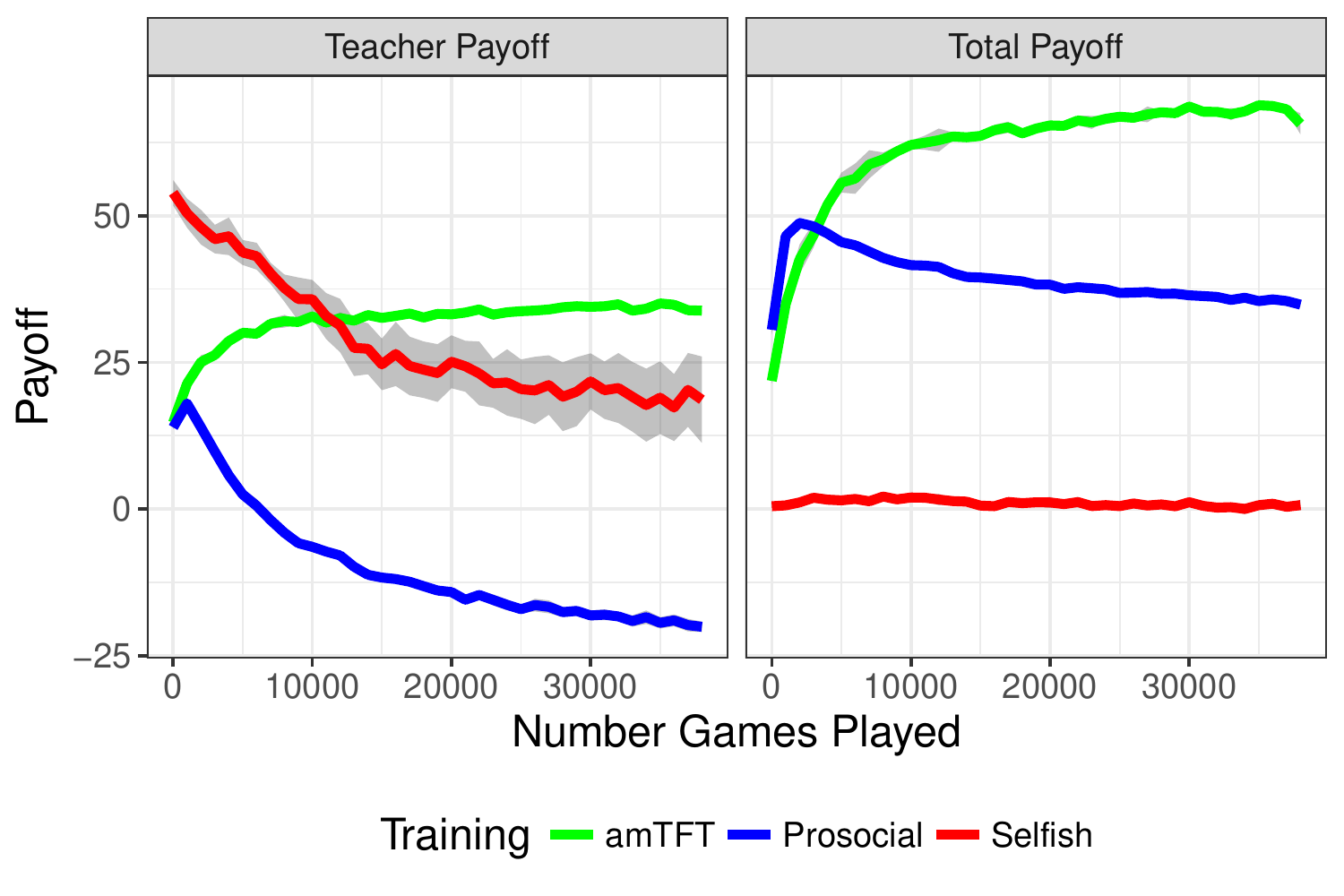}
\caption{Both purely selfish and purely cooperative Teachers lead Learners to exploitative strategies. However, amTFT Teachers lead Learners to cooperate and thus both agents reach a higher payoff in the long-run.}
\label{retraining_fig}
\end{figure}

\section{Conclusion}
Humans are remarkably adapted to solving bilateral social dilemmas. We have focused on how to give artificial agents this capability. We have shown that amTFT can maintain cooperation and avoid exploitation in Markov games. In addition we have provided a simple construction for this strategy that requires no more than modified self-play. Thus, amTFT can be applied to social dilemmas in many environments.

Our results emphasize the importance of treating agents as fundamentally different than other parts of the environment. In particular, agents have beliefs, desires, learn, and use some form of optimization while objects follow simple fixed rules. An important future direction for constructing cooperative agents is to continue to incorporate ideas from inverse reinforcement learning \citep{abbeel2004apprenticeship,ng2000algorithms} and cognitive science \citep{baker2009action,weiner2016coordinate} to construct agents that exhibit some theory of mind.

There is a growing literature on hybrid systems which include both human and artificial agents \citep{crandall2017cooperating, shirado2017locally}. In this work we have focused on defining `cooperation' as maximizing the joint payoff. This assumption seems reasonable in symmetric situations such as those we have considered, however, as we discuss in the introduction it may not always be appropriate. The amTFT construction can be easily modified to allow other types of focal points simply by changing the modified reward function used in the training of the cooperative strategies (for example by using the inequity averse utility functions of \cite{fehr1999theory}). However moving forward in constructing agents that can interact in social dilemmas with humans will require AI designers (and their agents) to understand and adapt to human cooperative and moral intutions \citep{yoeli2013powering,hauser2014cooperating,ouss2015punishment}. 

\bibliography{20180202_amtft.bbl}
%\bibliography{mtft_bib}
\bibliographystyle{icml2018}

\newpage

\section{Appendix}
\subsection{Standard Self-Play Fails to Discover Cooperative Strategies in the Repeated PD}
In a social dilemma there exists an equilibrium of mutual defection, and there may exist additional equilibria of conditional cooperation. Standard self-play may converge to any of these equilibria. When policy spaces are large, it is often the case that simple equilibria of constant mutual defection have larger basins of attraction than policies which maintain cooperation.

We can illustrate this with the simple example of the repeated Prisoner's Dilemma. Consider a PD with payoffs of $0$ to mutual defection, $1$ for mutual cooperation, $w > 1$ for defecting on a cooperative partner and $-s$ for being defected on while cooperating. Consider the simplest possible state representation where the set of states is the pair of actions played last period and let the initial state be $(C,C)$ (this is the most optimistic possible setup). We consider RL agents that use policy gradient (results displayed here come from using Adam \citep{kingma2014adam}, similar results were obtained with SGD though convergence speed was much more sensitive to the setting of the learning rate) to learn policies from states (last period actions) to behavior. 

Note that this policy space contains TFT (cooperate after $(C,C), (D,C)$, defect otherwise), Grim Trigger (cooperate after $(C,C)$, defect otherwise) and Pavlov or Win-Stay-Lose-Shift (cooperate after $(C,C), (D,D)$, defect otherwise \citep{nowak1993strategy}) which are all cooperation maintaining strategies (though only Grim and WSLS are themselves full equilibria).

Each episode is defined as one repeated PD game which lasts a random number of periods with stopping probability of stopping $.05$ after each period. Policies in the game are maps from the one-memory state space $\lbrace (C,C), (D,C), (C,D), (D,D) \rbrace$ to either cooperation or not. These policies are trained using policy gradient and the REINFORCE algorithm \citep{williams1992simple}. We vary $w$ and set $s = 1.5 w$ such that $(C,C)$ is the most efficient strategy always. Note that all of these parameters are well within the range where humans discover cooperative strategies in experimental applications of the repeated PD \citep{bo2011evolution}.

\begin{figure*}[h]
\centering
\includegraphics[scale=.7]{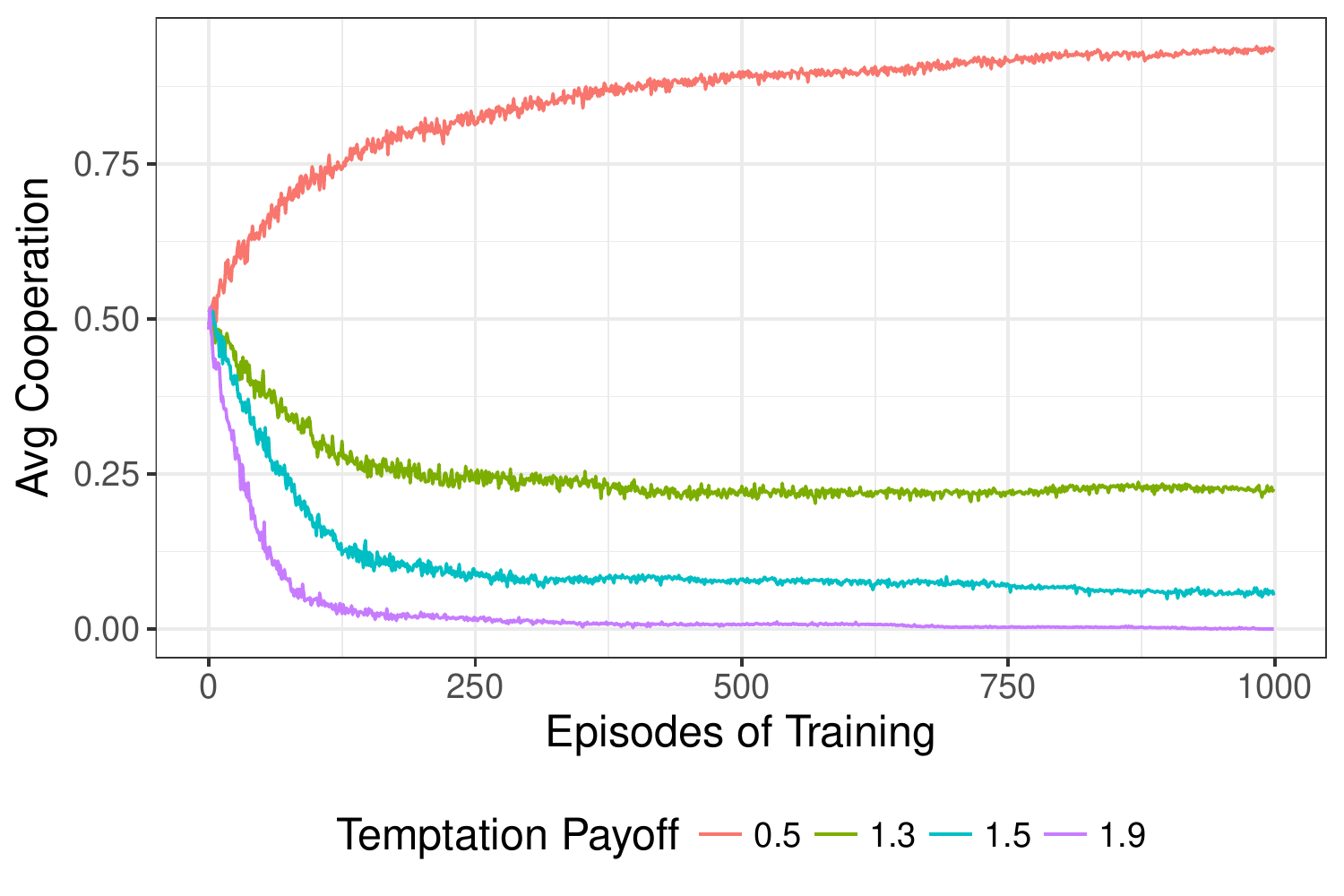}
\caption{Results from training one-memory strategies using policy gradient in the repeated Prisoner's Dilemma. Even in extremely favorable conditions self-play fails to discover cooperation maintaining strategies. Note that temptation payoff $.5$ is not a PD and here $C$ is a dominant strategy in the stage game.}
\label{pd_pg_results}
\end{figure*}

Figure \ref{pd_pg_results} shows that cooperation only robustly occurs when it is a dominant strategy for both players ($w < 0$) and thus the game is no longer a social dilemma.\footnote{Note that these results use pairwise learning and therefore are different from evolutionary game theoretic results on the emergence of cooperation \citep{nowak2006evolutionary}. Those results show that indeed cooperation can robustly emerge in these kinds of strategy spaces under evolutionary processes. Those results differ because they rely on the following argument: suppose we have a population of defectors. This can be invaded by mutants of TFT because TFT can try cooperation in the first round. If it is matched with a defector, it loses once but it then defects for the rest of the time, if it is matched with another TFT then they cooperate for a long time. Thus, for sufficiently long games the risk of one round of loss is far smaller than the potential fitness gain of meeting another mutant. Thus TFT can eventually gain a foothold. It is clear why in learning scenarios such arguments cannot apply.}. 

\subsection{Proof of Main Theorem}
To prove the theorem we will apply the one deviation principle. To show this, we fix player $1$ to be an amTFT agent and look at player $2$. Note that from the point of view of player $2$ this is now a Markov game with a state representation of $(s, k)$ where if $k=0$ player $1$ behaves according to $\pi^C$ and if $k>0$ player $1$ is in the $D$ phase and thus behaves according to $\pi^{D_k C}$. 

We consider the policy for player $2$ of `play $\pi^C_2$ when player $1$ is in the $C$ phase and play $\pi^D_2$ when player $1$ is in the $D$ phase.' Recall by the Principle of Optimality if there does not exist a one shot deviation $a'$ at any state under which player $2$ earns a higher payoff, then there does not exist a better policy than the one prescribed.

Consider starting at $k>0$. The suggested policy has player $2$ play $\pi^{D_k C}_2$. By $\pi^D$-dominance this is the best response to  $\pi^{D_k C}_1$ so there are no one-shot deviations in the $D$ phase.

Let us consider what happens in the $C$ phase ($k=0$). By the assumption of the theorem at any state $s$ we know that $$V_2 (s, \pi_1^C, \pi_2^C) - V_2 (s, \pi_1^D, \pi_2^D) > \frac{d^*}{\delta}.$$ Let $\lbrace r^t_{2} (s, \pi_1, \pi_2) \rbrace$ be the per-period reward stream (note here each $r$ is a random variable) for player $2$ induced by the policies $\pi_1, \pi_2.$ Since $$V_2 (s, \pi_1, \pi_2) = \mathbb{E} \left[ \delta^t \sum_{t=0}^{\infty} r^t_{2} (s, \pi_1, \pi_2) \right]$$  where $\delta$ is the discount rate. Because rewards are bounded then for any $\epsilon > 0$ there exists $k$ such that $$| V_2 (s, \pi_1, \pi_2) - \sum_{t=0}^{k} \delta^t r^t_{2} | < \epsilon.$$ That is, the first $k$ terms time steps approximate the full discounted expectation arbitrarily well. This also means that for some $k$ $$V_2 (s, \pi_1^C, \pi_2^C) - V_2 (s, \pi_1^{D_k C}, \pi_2^{D_k C}) > \frac{d^*}{\delta}.$$ From any state, the highest profit an agent can make from deviating from $\pi^C_2$ with a single action with an amTFT partner is $d^*.$ However we have shown that there exists a length $k$ such that moving to $D$ for $k$ turns costs the agent more than $\frac{d^*}{\delta}.$ Therefore there is no time during the $C$ phase they wish to deviate. This completes the proof.

\subsection{Aside on $\pi^{D}$-Dominance}
The reason we made the $\pi^{D}$-dominance assumption is to bound the expected payoff of an agent playing against $\pi^{D_k C}$ and therefore bound the necessary length of a $D$ phase after a particular deviation. However, in order to compute what the length of the $D$ phase the amTFT agent needs access to the best response policy to $\pi^{D_k C}$, or its associated value function. With $\pi^{D}$-dominance we assume that $\pi^D$ is that best response. Even if $\pi^{D}$-dominance does not strictly hold, it is likely a sufficient approximation. If necessary however, one can train an RL agent on episodes where their partner plays $\pi^{D_k C}$, where $k$ is observed. This allows one to approximate the best response policy to $\pi^{D_k C}$ which will then give us what we need to compute the responses to deviations from $\pi^C$ in the $D$ phase that incentivize full cooperation.

\subsection{Experimental Details}
We used rollouts to calculate the debit to the amTFT's partner at each time period. This estimator has good performance for both PPD and Coins given their reward structure. It is also possible to use a learned model of $Q$. Learning a sufficiently accurate model $\hat{Q}$ is challenging for several reasons. First, it has to have very low bias, since any bias in $\hat{Q}$ will be accumulated over periods. Second, the one-shot deviation principle demands that $\hat{Q}$ be accurate for all state-action pairs, not just those sampled by the policies $(\pi^C, \pi^C)$. Standard on-policy value function estimation will only produce accurate estimates of $Q$ at states sampled by the cooperative policies. As an example, in Coins, since the cooperative policies never collect their partner's coins $\hat{Q}$ for these state-action pairs may be inaccurate.

We found that it was possible in Coins to learn a model $\hat{Q}$ to calculate debit without policy rollouts using the same neural network architecture that was used to train the policies. However, we found that in order to train a $\hat{Q}$ model accurate enough to work well we had to use a modified training procedure. After finishing Selfish and Cooperative training, we perform a second step of training using a fixed (converged) $\hat{\pi}^C$. In order to sample states off the path of $\hat{\pi}^C$ during this step, the learner behaves according to a mixture of $\pi^C$, $\pi^D$, and random policies while the partner continues according to $\hat{\pi}^C$. $\hat{Q}$ is updated via off-policy Bellman iteration. We found this modified procedure produced a $\hat{Q}$ function that was good enough to maintain cooperation (though still not as efficient as rollouts). For more complex games, an important area for future work is to develop methodologies to compute more accurate approximations of $Q$ or combine a $\hat{Q}$ model with rollouts effectively.

\subsubsection{Coins Game and Training}
For Coins there are four actions (up, down, left, right), and $S$ is represented as a $4 \times 5 \times 5$ binary tensor where the first two channels encode the location of the each agent and the other two channels encode the location of the coin (if any exist). At each time step if there is no coin on the board a coin is generated at a random location with a random color, with probability $0.1$.

A policy $\pi(s;\theta): s \to \Delta(a)$ is learned via the advantage actor critic algorithm. We use a multi-layer convolutional neural network to jointly approximate the policy $\pi$ and state-value function $\hat{V}$. For this small game, a simpler model could be used, but this model generalizes directly to games with higher-dimensional 2D state spaces (e.g. environments with obstacles). For a given board size $k$, the model has $\lceil \log_2(k) \rceil + 1$ repeated layers, each consisting of a 2D convolution with kernel size 3, followed by batch normalization and ReLU. The first layer has stride 1, while the successive layers each have stride 2, which decreases the width and height from $k$ to $\lceil k/2 \rceil$ while doubling the number of channels. For the $5 \times 5$ board, channel sizes are 13, 26, 52, 104. From these 104 features, $\pi$ is computed via a linear layer with 4 outputs with softmax, to compute a distribution over actions, while the value function is computed via a single-output linear layer.

The actor and critic are updated episodically with a common learning rate - at the end of each game we update the model on a batch of episodes via $$\Delta \theta_i = \lambda \left( A_t \frac{\partial V(s_t)}{\partial \theta_i} + \tilde{A_t} \log\pi(s_t, a_t) \frac{\partial \pi(s_t, a_t)}{\partial \theta_i} \right)$$

where $A$ is the advantage $$A_t = r_t + \delta V(s_{t+1}) - V(s_t)$$

and $\tilde{A}$ is the advantaged normalized over all episodes and periods in the batch

\begin{equation*}
\tilde{A}_t = \frac{A_t - |A|}{\sigma(A)} .
\end{equation*}

We train with a learning rate of $0.001$, continuation probability $.998$ (i.e. games last on average 500 steps), discount rate $0.98$, and a batch size of $32$. We train for a total of $40,000$ games.

\subsubsection{Pong Player Dilemma Training}

We use the arcade learning environment modified for 2-player play as proposed in \cite{tampuu2017multiagent}, with modified rewards of +1 for scoring a point and -2 for being scored on. We train policies directly from pixels, using the pytorch-a3c package \url{https://github.com/ikostrikov/pytorch-a3c}. 

Policies are trained directly from pixels via A3C \citep{mnih2016asynchronous}. Inputs are rescaled to 42x42 and normalized, and we augment the state with the difference between successive frames with a frame skip of $8$. We use 38 threads for A3C, over a total of 38,000 games (1,000 per thread). We use the default settings from pytorch-a3c: a discount rate of $0.99$, learning rate of $0.0001$, 20-step returns, and entropy regularization weight of $0.01$.

The policy is implemented as a convolutional neural network with four layers, following pytorch-a3c. Each layer uses a 3x3 kernel with stride 2, followed by ELU. The network has two heads for the actor and critic. We elide the LSTM layer used in the pytorch-a3c library, as we found it to be unnecessary.

\subsubsection{Tournament Results}

\begin{figure*}[!htb]
  \centering
    \subfloat[Coins Results]{{\includegraphics[scale=.51]{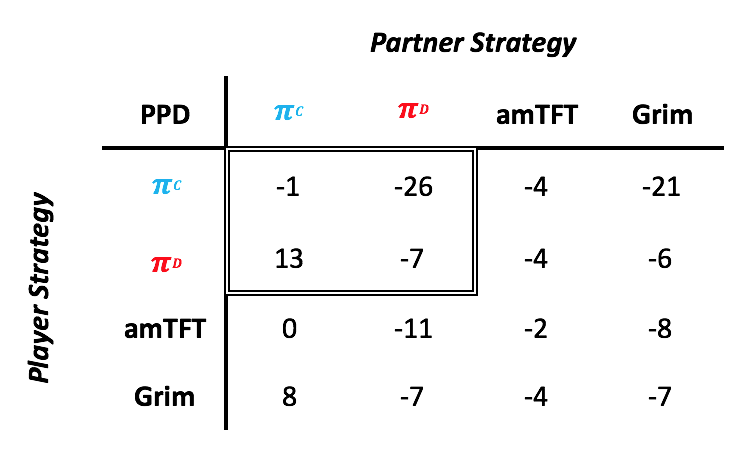}}} 
  \subfloat[PPD Results]{{\includegraphics[scale=.51]{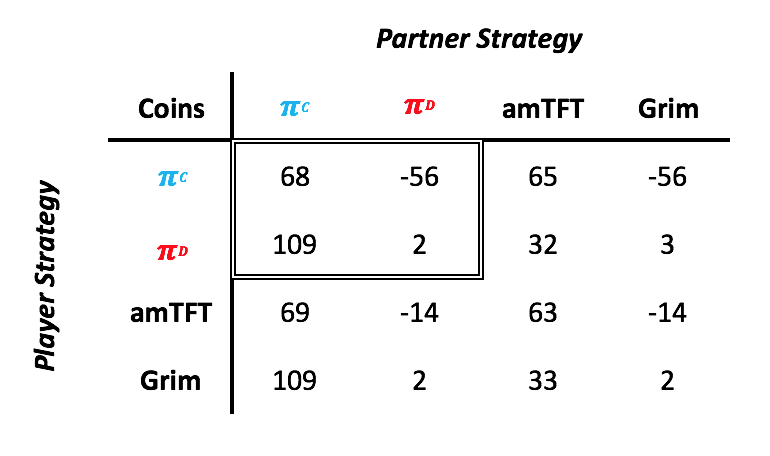}}}

  \caption{Results of the tournament in two Markov social dilemmas. Each cell contains the average total reward of the row strategy against the column strategy. amTFT achieves close to cooperative payoffs with itself and achieves close to the defect payoff against defectors. Its partner also receives a higher payoff for cooperation than defection.}
  \label{expt_results}
\end{figure*}

\end{document}